\documentclass[twoside,11pt]{article}

\usepackage{jmlr2e}

\usepackage{mathtools}
\usepackage[T1]{fontenc}
\usepackage{lmodern}
\usepackage{enumitem}
\usepackage{lastpage}
\input{glyphtounicode}\pdfgentounicode=1 

\newtheorem{assumption}[theorem]{Assumption}

\DeclareMathOperator*{\argmax}{arg\,max}

\DeclareMathOperator{\supp}{supp}

\DeclareMathOperator{\E}{\mathbb{E}}
\DeclareMathOperator{\Rad}{\mathfrak{R}}
\newcommand{\R}{\mathbb{R}}

\newcommand{\Sstar}{\Sigma^{\!*}}
\newcommand{\indic}{\mathbf{1}}
\newcommand{\meanfloor}{\bar\ell_0}
\newcommand{\maxfloor}{\ell_0^{\max}}
\newcommand{\Bp}{\bar B'}

\ShortHeadings{A Reverse-Gold Theorem}{Sengupta}
\firstpageno{1}

\begin{document}

\title{A Reverse-Gold Theorem:
	Positive-Data Identification by Substrate-Bounded Contrastive Learners}

\author{\name Rakesh Sengupta \email qg.rakesh@gmail.com \\
       \addr School of Interwoven Arts and Sciences\\
       Krea University\\
       Sri City, AP 517646, India
      }

\editor{}

\maketitle

\begin{abstract}
Gold's theorem shows that no learner can identify an arbitrary language in a superfinite class from positive presentations alone. We prove a positive result of a different shape. Fix an innate \emph{recursive contrast operator} $R$ mapping any string to a finite set of structurally bounded perturbations of it, and a discriminative scorer of bounded capacity trained by InfoNCE-style contrastive estimation \citep{vandenoord2018,macollins2018} on self-generated perturbations. With high probability over the sample, the trained scorer's \emph{misranking probability}---the chance that a positive scores no higher than some out-of-language perturbation in its own orbit---is at most an explicit approximation-gap term plus $\varepsilon$, with sample complexity polynomial in the capacity, the orbit size on the support of the input distribution, the score bound, $1/\varepsilon$, and an inverse loss margin fixed by the score bound and the orbit size. When the substrate saturatedly realizes the separation and the operator is orbit-pure, the gap vanishes and the scorer achieves \emph{PAC conditional identification} from positive data alone: it ranks all but an $\varepsilon$-fraction of positives above every out-of-language perturbation in their own orbit. The negatives are constructed by the learner, not supplied by an oracle, and no global normalizing constant over strings is ever computed. The misranking bound follows from the geometry of the contrastive loss---per-example loss floors plus a definite loss increment under misranking---yielding an $\varepsilon^{-2}$ rate, and, under saturated realizability, the approximation gap is at most the expected log-number of in-language (``colliding'') perturbations. The ranking guarantee lifts to a global membership classifier under a margin condition, and the same operator composed with the learned scorer is a normalization-free Metropolis--Hastings generator. We verify the hypotheses end to end on a strictly local language, sketch correspondences with substitutable context-free languages \citep{clarkeyraud2007} and the subregular classes \citep{heinz2010,hayashikamigaito2025}, and state an open conjecture locating the construction in the mildly context-sensitive hierarchy.
\end{abstract}


\section{Introduction}

	\citet{gold1967} proved that for any class $\mathcal{C}$ of languages containing all finite languages and at least one infinite language, no learner identifies an arbitrary $L\in\mathcal{C}$ in the limit from a positive presentation. The proof is adversarial: a teacher can withhold the strings that would distinguish a finite hypothesis from an infinite target, and a positive-only learner cannot falsify the finite hypothesis without seeing strings known to be absent from $L$. Since the context-free and context-sensitive classes are superfinite, neither is Gold-identifiable from positive data. The usual conclusions are that either explicit negative evidence is available, or the hypothesis space must be restricted in advance, or the identification criterion must be weakened from exact convergence to approximate (PAC) convergence \citep{valiant1984}.
	
	This paper isolates a fourth structural ingredient that makes positive-data learning go through without external negatives and without a stipulated restriction on \emph{which} languages are possible. Suppose the learner is equipped with a fixed \emph{recursive contrast operator} $R:\Sstar\to 2^{\Sstar}$ that maps any string to a finite set of structurally bounded perturbations of it---substitution of a like-typed constituent, deletion, local reordering, and compositions of these---defined uniformly over all strings without reference to the target. Then, given a positive example $x$, the learner can manufacture its own candidate negatives $R(x)$, without an oracle, even though it does not know which elements of $R(x)$ are in fact out-of-language. Pairing $R$ with a bounded-capacity discriminative scorer and training by a contrastive (InfoNCE) objective, we ask: can the learner identify the target from positive data alone, using its self-generated perturbations as implicit negatives?
	
	We answer with a weaker but precise positive result: a \emph{conditional-ranking} guarantee. With high probability, the trained scorer's misranking probability is at most an explicit approximation-gap term $a_{\mathcal{H}}/(\log 2-\maxfloor)$ plus $\varepsilon$, with sample complexity polynomial in the capacity, the orbit size on the support, the score bound, $1/\varepsilon$, and the inverse loss margin $(\log 2-\maxfloor)^{-1}$. When the gap vanishes---under saturated realizability and an orbit-pure operator---the scorer ranks all but an $\varepsilon$-fraction of positives drawn from $p_d$ strictly above every out-of-language string in their own orbit. This is the honest primitive that the contrastive objective delivers; we are careful to distinguish it from a global membership classifier, which we recover only under an additional margin assumption (Corollary~\ref{cor:classification}), and we quantify, rather than assume away, the cost of perturbations that land \emph{inside} the target---the contrastive analogue of class collision \citep{arora2019}---through an explicit approximation gap (Proposition~\ref{prop:collision}).
	
	\paragraph{Contributions.}
	\begin{enumerate}[leftmargin=2em]
		\item We formalize the recursive contrast operator and a faithfulness condition (boundary support plus realizability) relating the operator, the bounded-capacity scorer, and the target. We prove an agnostic misranking bound for the InfoNCE-trained scorer: an explicit approximation-gap term $a_{\mathcal{H}}/(\log2-\maxfloor)$ plus $\varepsilon$ (Theorem~\ref{thm:main}), and show that the gap vanishes under saturated realizability and an orbit-pure operator, which yields PAC conditional identification from positive data (Proposition~\ref{prop:collision}).
		\item The misranking bound is derived from the loss geometry directly. The per-example InfoNCE loss has a known per-example floor, and any misranking forces the loss above $\log 2$; together these give a \emph{linear} relationship between misranking probability and excess loss, hence an $\varepsilon^{-2}$ sample complexity, with the approximation error appearing as an explicit additive term (Lemma~\ref{lem:cal}). This avoids routing through a generic surrogate-calibration transform, which would only give $\varepsilon^{-4}$.
		\item We analyse \emph{orbit collisions}---perturbations of a positive that remain in-language. Under a saturated form of realizability the approximation gap is bounded by the expected log-collision count $\E[\log(1+m^+(x))]$, which vanishes for orbit-pure operators; and when, with positive probability, an orbit contains a support point that has a non-empty orbit of its own, the gap is strictly positive for any scorer, an information-theoretic feature of self-supervision rather than proof slack (Proposition~\ref{prop:collision} and Remark~\ref{rem:intrinsic}).
		\item We track the dependence on orbit size correctly through a vector-contraction step \citep{maurer2016}, which the loss-Lipschitz heuristic obscures (Lemma~\ref{lem:uc}).
		\item We show that the same operator $R$ composed with the learned scorer is a normalization-free Metropolis--Hastings sampler whose stationary law, on finite components, is the score-reweighted measure on the seed's component under the symmetrized operator (Proposition~\ref{prop:generator}); the construction is thus a controlled generator as well as a discriminator.
		\item We exhibit an explicit end-to-end instance with zero approximation gap on a strictly local language (Appendix~\ref{app:example}), sketch how the framework corresponds to the classical positive-data results for substitutable context-free languages and the subregular classes (Section~\ref{sec:instances}), and state a conjecture locating the construction in the mildly context-sensitive hierarchy (Section~\ref{sec:open}).
	\end{enumerate}
	
	\paragraph{Scope.} This is a formal learnability paper. The construction was motivated by a question in the theory of language acquisition---whether the poverty-of-the-stimulus argument's appeal to innate grammatical content can be replaced by an innate \emph{procedural} operator plus a capacity bound---but we develop only the mathematics here and state the cognitive interpretation separately. Where a term has a linguistic motivation (``recursive,'' ``constituent,'' ``closure''), it is given a precise formal meaning and used only in that sense.
	
	\section{Formal Setup}\label{sec:setup}
	
	Let $\Sigma$ be a finite alphabet and $\Sstar$ the set of finite strings over $\Sigma$. A \emph{language} is a subset $L \subseteq \Sstar$.
	
	\begin{definition}[Input distribution]
		The learner receives a sample $D = \{x_1, \ldots, x_n\}$ drawn i.i.d.\ from a distribution $p_d$ on $\Sstar$ with $\supp(p_d) \subseteq L$. We do not assume $\supp(p_d) = L$; the input distribution may concentrate on a sub-region of the target.
	\end{definition}
	
	\begin{definition}[Recursive contrast operator]\label{def:R}
		A \emph{recursive contrast operator} on $\Sstar$ is a map $R: \Sstar \to 2^{\Sstar}$ assigning to each $x$ a finite set $R(x) \subseteq \Sstar$ of perturbations, satisfying:
		\begin{enumerate}[leftmargin=2em]
			\item \emph{Locality.} Each $x' \in R(x)$ differs from $x$ by a structurally bounded perturbation: substitution of a constituent by a like-typed constituent, deletion, insertion, reordering, or composition of such operations up to bounded depth, defined over the constituent bracketing of $x$ (Remark~\ref{rem:constituency}).
			\item \emph{Universality.} $R$ is defined uniformly over $\Sstar$ without reference to any specific target $L$.
			\item \emph{Recursivity.} $R$ is compatible with structural composition: if $x = uv$ with $u, v$ in the relevant structural class, then $R(x) \supseteq \{u'v : u' \in R(u)\} \cup \{uv' : v' \in R(v)\}$.
		\end{enumerate}
		We write $m(x) := |R(x)|$, which is finite for every $x$. The uniform orbit bound is needed only on the inputs the learner sees: we write $M := \sup_{x\in\supp(p_d)} m(x)$ and assume $M<\infty$. A global bound $\sup_{x\in\Sstar} m(x)<\infty$ is \emph{not} assumed, and it typically fails for natural operators; for the operator of Example~\ref{ex:sl2}, $m((ab)^n)=n$. Every use of $M$ below is either at sampled points, which lie in $\supp(p_d)$ almost surely, or inside an expectation over $p_d$.
	\end{definition}
	
	The canonical instance for context-free targets is substitution of distributionally equivalent constituents, which underpins the algorithm of \citet{clarkeyraud2007}; richer operators (adjoining, multi-component substitution) are required higher in the hierarchy, as discussed in Section~\ref{sec:open}.
	
	\begin{remark}[Constituency as procedure versus output]\label{rem:constituency}
		The locality clause is defined over constituents, and a natural objection is that this presupposes the bracketing that learning is meant to deliver. The resolution is to read $R$ as a \emph{procedure}---``substitute a like-typed constituent,'' etc.---defined as a function of whatever bracketing is currently hypothesized, while $R(x)$ as an \emph{output} depends on the current state of learning. Started from a trivial (flat) bracketing, $R$ degenerates to surface substitution; the contrastive signal under that degenerate $R$ is intended to separate substitutable from non-substitutable spans, in the distributional sense that defines constituents \citep{clarkeyraud2007}; the induced bracketing refines the input to $R$ on the next pass. Bracketing and scorer are then recovered together as a joint fixed point. The theorems below are stated for a \emph{fixed} operator. They apply to the operator induced by any fixed bracketing, including one estimated on data independent of the training sample. They do not analyse the iterative procedure in which the bracketing is re-estimated from the same sample, because the orbits then depend on the data; sample splitting across passes would restore the independence each pass requires. We leave the joint fixed point as a procedure, not a theorem.
	\end{remark}
	
	\begin{definition}[Discriminative hypothesis class]\label{def:hclass}
		A \emph{discriminative hypothesis class} of capacity $\kappa$ is a parametrized family $\mathcal{H} = \{s_\theta : \Sstar \to \R \mid \theta \in \Theta\}$ of scoring functions whose empirical Rademacher complexity satisfies $\widehat{\Rad}_S(\mathcal{H}) \leq \sqrt{\kappa/n}$ for \emph{every} $S\in\mathcal{X}_R^{\,n}$ and all $n$, where $\mathcal{X}_R := \supp(p_d)\cup\bigcup_{x\in\supp(p_d)}R(x)$. Since $\mathcal{X}_R$ depends on $R$ and $p_d$, so in general does $\kappa$. The sample-uniform (worst-case) form of the bound is used in Lemma~\ref{lem:uc}, where the class is evaluated on deterministically transformed copies of the data; those copies lie in $\mathcal{X}_R$, which is all the lemma needs. We use Rademacher complexity rather than VC dimension because the scorers are real-valued; for the combinatorial instances below the two are interchangeable up to logarithmic factors via Dudley/VC chaining \citep{mohri2018}.
	\end{definition}
	
	\begin{assumption}[Bounded scores]\label{ass:bounded}
		There is a constant $B > 0$ with $|s_\theta(x)| \leq B$ for every $\theta \in \Theta$ and every $x \in \Sstar$.
	\end{assumption}
	
	Assumption~\ref{ass:bounded} holds for the architectures we have in mind---bounded-weight recurrent networks, transformers with saturating readouts, finite-feature linear scorers---with $B$ an architectural constant fixed by the substrate, not scaling with the input. Without a score bound the InfoNCE log-loss is unbounded below and uniform-convergence arguments do not apply directly.
	
	\begin{definition}[Pointwise faithfulness]\label{def:faithful}
		Fix a language $L$ and an input distribution $p_d$ supported on $L$. The pair $(R,\mathcal{H})$ is \emph{faithful to $L$} (relative to $p_d$) if:
		\begin{enumerate}[leftmargin=2em]
			\item \emph{Boundary support (a.e.).} For $p_d$-almost every $x \in L$, the orbit $R(x)$ contains at least one element of $\Sstar \setminus L$.
			\item \emph{Realizability.} There is $\theta^* \in \Theta$ with $s_{\theta^*}(x) > s_{\theta^*}(x')$ for $p_d$-almost every $x \in L$ and every $x' \in R(x) \setminus L$.
		\end{enumerate}
	\end{definition}
	
	The two clauses are logically independent: boundary support is a property of $R$ and $L$ (does the operator expose the boundary?); realizability is a property of $\mathcal{H}$ and $L$ (can the scorer represent it once exposed?). As Section~\ref{sec:bias} and Theorem~\ref{thm:main} make precise, neither clause is needed for the misranking \emph{inequality}---which is agnostic---but they interpret it: boundary support makes the guarantee non-vacuous, and realizability makes the problem well-posed. Realizability alone does not make the approximation gap small; its saturated form (Definition~\ref{def:saturated}) bounds the gap by the collision term, which vanishes for orbit-pure operators (Proposition~\ref{prop:collision}). Boundary support is stated ``almost everywhere'' deliberately: strings whose orbit happens to lie inside $L$ contribute no contrastive signal but do not obstruct the proof.
	
	\begin{definition}[Saturated realizability]\label{def:saturated}
		The pair $(R,\mathcal{H})$ is \emph{saturatedly realizable for $L$} (relative to $p_d$) if there is $\theta^\star \in \Theta$ with $s_{\theta^\star}(x) = B$ for $p_d$-almost every $x \in L$ and $s_{\theta^\star}(x') = -B$ for every $x' \in R(x)\setminus L$. Saturated realizability implies clause~2 of Definition~\ref{def:faithful}; it additionally requires that the realizing scorer can be driven to the score bound. A sufficient condition is a realizer with a \emph{uniform margin about a common threshold}---some $\theta$, $\tau$ and $\gamma>0$ with $s_\theta(x)\ge\tau+\gamma$ for $p_d$-a.e.\ $x\in L$ and $s_\theta(x')\le\tau-\gamma$ for every $x'\in R(x)\setminus L$---together with closure of $\mathcal{H}$ under the maps $s\mapsto\mathrm{clip}_{[-B,B]}\big(c\,(s-\tau)\big)$ for $c\ge B/\gamma$. A strict realizer alone does not suffice: it ranks within each orbit, but score ranges of different orbits may overlap (Remark~\ref{rem:rankingform}), so no single rescaling sends every positive to $B$.
	\end{definition}
	
	Note that Definition~\ref{def:saturated} constrains $\theta^\star$ only on positives and on \emph{out-of-language} orbit elements; in-language orbit elements $x'\in R(x)\cap L$ may score anywhere in $[-B,B]$, and indeed must score $+B$ if they also lie in $\supp(p_d)$. Proposition~\ref{prop:collision} quantifies the cost of these \emph{collisions}.
	
	\begin{definition}[Uniform faithfulness]\label{def:uniform}
		A single pair $(R,\mathcal{H})$ is \emph{uniformly faithful to a class $\mathcal{C}$} if it is faithful to every $L\in\mathcal{C}$ (under each admissible $p_d$). Theorem~\ref{thm:main} is proved one target at a time and uses only pointwise faithfulness of the $L$ being learned. Uniform faithfulness---one operator and one substrate adequate across the entire class---is the substantially stronger property: it is what the conjecture of Section~\ref{sec:open} bets on, and establishing it for a mildly context-sensitive class is itself a major open problem, since it quantifies over every language in the class at once.
	\end{definition}
	
	\begin{definition}[InfoNCE objective]\label{def:infonce}
		For $\theta\in\Theta$, the InfoNCE objective \citep{vandenoord2018,macollins2018} is
		\begin{equation}\label{eq:infonce}
			J(\theta) = \E_{x \sim p_d} \left[ \log \frac{\exp s_\theta(x)}{\exp s_\theta(x) + \sum_{x' \in R(x)} \exp s_\theta(x')} \right],
		\end{equation}
		with empirical version $\hat{J}_n$ replacing the expectation by the average over $D$. Equivalently, $-J(\theta)$ is the expected multinomial-logistic loss of the $(m(x){+}1)$-way classification problem ``identify $x$ among $\{x\}\cup R(x)$,'' with $x$ the correct class.
	\end{definition}
	
	\section{Main Result}\label{sec:theorem}
	
	Throughout, write
	\[
	\ell_0(x) \;:=\; \log\!\big(1 + m(x)\, e^{-2B}\big), \qquad
	\meanfloor \;:=\; \E_{x\sim p_d}\big[\ell_0(x)\big], \qquad
	\maxfloor \;:=\; \log\!\big(1 + M e^{-2B}\big)
	\]
	for the \emph{per-example} loss floor identified in Lemma~\ref{lem:range}, its mean under the input distribution, and its worst case over the support ($\maxfloor = \sup_{x\in\supp(p_d)} \ell_0(x)$). Define the \emph{approximation gap}
	\[
	a_{\mathcal{H}} \;:=\; \Big(\inf_{\theta\in\Theta} -J(\theta)\Big) - \meanfloor \;\geq\; 0,
	\]
	the amount by which the best scorer in $\mathcal{H}$ exceeds the mean floor; nonnegativity is part of Lemma~\ref{lem:range}. When orbit sizes vary, the per-example floor---not the worst-case floor---is the correct baseline: it is what makes the pointwise calibration of Lemma~\ref{lem:cal} valid, and what makes $a_{\mathcal{H}}$ attain zero in the orbit-pure case under saturated realizability (Proposition~\ref{prop:collision}). Section~\ref{sec:bias} interprets $a_{\mathcal{H}}$.
	
	\begin{theorem}[Reverse-Gold, ranking form]\label{thm:main}
		Fix any language $L$, operator $R$ with orbit bound $M$ on the support (Definition~\ref{def:R}), class $\mathcal{H}$ of capacity $\kappa$ and score bound $B$ with $B > \tfrac12\log M$, and input distribution $p_d$ supported on $L$. Then for every $\varepsilon, \delta \in (0,1)$, the InfoNCE-trained scorer $s_{\hat\theta_n}$ with $\hat\theta_n = \argmax_\theta \hat J_n(\theta)$ (for a $\rho$-approximate maximizer, replace $\varepsilon$ by $\varepsilon+\rho/(\log2-\maxfloor)$; Remark~\ref{rem:interpretation}) obtained from
		\[
		n \;=\; O\!\left( \frac{(M+1)^2\,\Bp^{\,2}\,(\kappa + \log(2/\delta))}{(\log 2 - \maxfloor)^2\,\varepsilon^2} \right), \qquad \Bp := \max\{1,\ 2B + \log(M+1)\},
		\]
		samples satisfies, with probability at least $1-\delta$,
		\[
		\Pr_{x \sim p_d}\!\Big[\, \exists\, x' \in R(x)\setminus L : \; s_{\hat\theta_n}(x) \leq s_{\hat\theta_n}(x') \,\Big]
		\;\leq\; \frac{a_{\mathcal{H}}}{\log 2 - \maxfloor} \;+\; \varepsilon .
		\]
	\end{theorem}
	
	\begin{remark}[Interpretation of the bound]\label{rem:interpretation}
		The bound is \emph{agnostic}: no faithfulness assumption is used in its derivation. The clauses of Definitions~\ref{def:faithful} and~\ref{def:saturated} interpret it rather than power it. Boundary support guarantees the misranking event is non-vacuous---some orbit elements lie outside $L$, so there is something to rank wrongly; absent it the left-hand side is trivially zero. Realizability makes the problem well-posed: $\mathcal{H}$ contains a correct-ranking scorer. Saturated realizability bounds the gap by the collision term, $a_{\mathcal{H}} \le \E[\log(1+m^+(x))]$ with $m^+(x) = |R(x)\cap L|$, and forces $a_{\mathcal{H}} = 0$ when the operator is \emph{orbit-pure} almost everywhere, $R(x)\cap L = \emptyset$ (Proposition~\ref{prop:collision}). In that case the conditional misranking probability is at most $\varepsilon$: the learned scorer ranks all but an $\varepsilon$-fraction of positives above every out-of-language string in their own orbit. We call this \emph{PAC conditional identification} of $L$ relative to $p_d$ and $R$. Realizability alone does not give it, and neither does saturated realizability without orbit purity (Example~\ref{ex:sl2bad}). If the empirical maximum is not attained, any $\rho$-approximate maximizer ($\hat J_n(\hat\theta_n)\ge\sup_\theta\hat J_n(\theta)-\rho$) satisfies the same bound with $\varepsilon$ replaced by $\varepsilon+\rho/(\log2-\maxfloor)$, by the same proof.
	\end{remark}
	
	The criterion is \emph{one-step} conditional ranking: each positive in $\supp(p_d)$ is ranked against its own orbit $R(x)$. Strings reachable from the support only in several steps---elements of the closure $\mathrm{Cl}_R(\supp(p_d))$, the smallest set containing $\supp(p_d)$ and closed under $R$---are never ranked by Theorem~\ref{thm:main}; they enter only through the separation condition of Corollary~\ref{cor:classification}. We are explicit (Section~\ref{sec:bias}) that this is a genuinely different criterion from Gold-style identification of $L$. The proof is assembled from four lemmas. The first (Lemma~\ref{lem:negative}) is bookkeeping; the analytic content is in Lemmas~\ref{lem:uc} and~\ref{lem:cal}, with the loss geometry of Lemma~\ref{lem:range} feeding both. No property of the population maximizer is needed. Proofs of Lemmas~\ref{lem:range} and~\ref{lem:uc} are deferred to Appendix~\ref{app:proofs}; the calibration argument, which carries the novelty, is given in the main text.
	
	\begin{lemma}[Self-negative construction]\label{lem:negative}
		For any $x$, the orbit $R(x)$ partitions into $R^+(x) = R(x) \cap L$ and $R^-(x) = R(x) \setminus L$, with $R^-(x) \neq \emptyset$ for $p_d$-almost every $x \in L$ by boundary support. Write $m^+(x) := |R^+(x)|$ and $m^-(x) := |R^-(x)|$. The learner has access to $x$ (known positive) and to $R(x)$ (elements of unknown status), but not to the partition. The boundary-support clause is used only to interpret Theorem~\ref{thm:main}, not in its proof.
	\end{lemma}
	
	\begin{proof}
		Immediate from Definitions~\ref{def:R} and~\ref{def:faithful}.
	\end{proof}
	
	The content of Lemma~\ref{lem:negative} is that the learner constructs candidate negatives without an oracle. Were the partition known, the learner would have explicit negatives and Gold's theorem would not bind; the difficulty---and the reason the contrastive route is non-trivial---is precisely that it is not known.
	
	InfoNCE, unlike vanilla noise-contrastive estimation \citep{gutmannhyvarinen2012}, does not require the noise distribution to dominate the data distribution: it estimates only the relative ranking of a positive against its own conditional negatives, which is all the construction uses, and remains well-defined even when $R(x)$ contains only out-of-language strings. Drawing negatives conditioned on the positive places the construction in the conditional-NCE family \citep{ceylangutmann2018}, of which $R$ is a structured, non-parametric instance: the conditional noise distribution is supported on the orbit of recursive perturbations rather than on a hand-chosen parametric family.
	
	\begin{lemma}[Loss range and per-example floor]\label{lem:range}
		The per-example loss $\ell(\theta;x) = -s_\theta(x) + \log\!\big(\exp s_\theta(x) + \sum_{x'\in R(x)} \exp s_\theta(x')\big)$ satisfies
		\[
		\ell(\theta;x) \;\in\; \big[\,\ell_0(x),\; 2B + \log(m(x)+1)\,\big] \;\subseteq\; \big[\,\ell_0(x),\; 2B + \log(M+1)\,\big]
		\]
		for every $\theta$ and $x$ (the second inclusion for $x\in\supp(p_d)$), where $\ell_0(x) = \log(1+m(x)e^{-2B})$. The floor $\ell_0(x)$ is the global minimum over score configurations in $[-B,B]^{m(x)+1}$ and, for $m(x)\ge1$, is attained only at $s_\theta(x)=B$ and $s_\theta(x')=-B$ for all $x'\in R(x)$. Consequently $-J(\theta) \ge \meanfloor$ for every $\theta$, so $a_{\mathcal{H}} \ge 0$. Note that $\ell_0(x) \le \maxfloor$ for $x\in\supp(p_d)$, with equality iff $m(x) = M$; when orbit sizes vary, the per-example loss can fall strictly below $\maxfloor$, which is why the floor must be indexed by $x$.
	\end{lemma}
	
	\begin{lemma}[Uniform convergence]\label{lem:uc}
		With probability at least $1-\delta$ over $D$,
		\[
		\big(-J(\hat\theta_n)\big) - \Big(\inf_{\theta} -J(\theta)\Big)
		\;\le\; C\,\frac{(M+1)\sqrt{\kappa} + (2B+\log(M+1))\sqrt{\log(2/\delta)}}{\sqrt n}
		\]
		for an absolute constant $C$.
	\end{lemma}
	
	\begin{remark}[On the $(M+1)$ factor]\label{rem:Mfactor}
		The orbit size enters Lemma~\ref{lem:uc} \emph{twice}: logarithmically through the loss range $B'=2B+\log(M+1)$, and linearly through the vector contraction, $(M+1)\sqrt{\kappa/n}$. The linear factor is the conservative cost of reusing one scorer across $M+1$ coordinates; the loss-Lipschitz heuristic (``softmax is $1$-Lipschitz'') hides it. It should be improvable: the $\ell_\infty$ refinement of the vector-contraction inequality and class-count-optimal multiclass bounds are expected to reduce it toward $\sqrt{M+1}$ (Section~\ref{sec:open}). We carry the conservative $(M+1)$ because the thesis needs only polynomiality, and state the bound honestly rather than tightly. In the worked instance of Appendix~\ref{app:example} the support orbit bound is at most linear in the maximal support length $K$ ($M\le\lceil K/2\rceil$), as it is for single-segment substitutions ($M\le(|\Sigma|-1)K$; $O(|\Sigma|K)$ with insertions and deletions), so the factor is polynomial; it bites hardest for operators of large fan-out, where the $\ell_\infty$ refinement becomes worth the extra work.
	\end{remark}
	
	\begin{lemma}[Direct calibration of misranking]\label{lem:cal}
		Assume $B > \tfrac12\log M$, so $\maxfloor < \log 2$. For every $\theta$,
		\[
		\Pr_{x\sim p_d}\big[\,\exists\, x'\in R(x)\setminus L:\ s_\theta(x)\le s_\theta(x')\,\big]
		\;\le\; \frac{\big(-J(\theta)\big) - \meanfloor}{\log 2 - \maxfloor}.
		\]
	\end{lemma}
	
	\begin{proof}
		Fix $x$ and suppose it is misranked: some $x'\in R(x)\setminus L$ has $s_\theta(x')\ge s_\theta(x)$. Then the denominator of \eqref{eq:infonce} is at least $e^{s_\theta(x)} + e^{s_\theta(x')} \ge 2 e^{s_\theta(x)}$, so the softmax mass on the positive is $p_0 \le \tfrac12$ and the per-example loss is $\ell(\theta;x) = -\log p_0 \ge \log 2$. By Lemma~\ref{lem:range}, $\ell(\theta;x)\ge \ell_0(x)$ always. Hence, pointwise for $x\in\supp(p_d)$, which carries all the mass of $p_d$,
		\[
		\indic[\text{$x$ misranked}] \;\le\; \frac{\ell(\theta;x) - \ell_0(x)}{\log 2 - \maxfloor},
		\]
		because when $x$ is misranked the numerator is $\ge \log 2 - \ell_0(x) \ge \log 2 - \maxfloor$ and the ratio is therefore $\ge 1$, and when $x$ is not misranked the right-hand side is $\ge 0 = $ the left-hand side (using $\ell\ge\ell_0(x)$ and $\log2>\maxfloor$). Taking expectations over $x\sim p_d$ gives $\Pr[\text{misrank}] \le (\E[\ell(\theta;x)]-\E[\ell_0(x)])/(\log2-\maxfloor) = ((-J(\theta))-\meanfloor)/(\log2-\maxfloor)$.
	\end{proof}
	
	\begin{proof}[Proof of Theorem~\ref{thm:main}]
		Apply Lemma~\ref{lem:cal} to $\hat\theta_n$:
		\[
		\Pr[\text{misrank under }\hat\theta_n] \;\le\; \frac{(-J(\hat\theta_n)) - \meanfloor}{\log2-\maxfloor}
		\;=\; \frac{a_{\mathcal{H}} + \big[(-J(\hat\theta_n)) - \inf_\theta(-J(\theta))\big]}{\log2-\maxfloor},
		\]
		using $\inf_\theta(-J(\theta)) = \meanfloor + a_{\mathcal{H}}$ by definition of the approximation gap. By Lemma~\ref{lem:uc} the bracketed estimation term is at most $C\big((M+1)\sqrt\kappa + B'\sqrt{\log(2/\delta)}\big)/\sqrt n \le \sqrt2\,C\,(M+1)\Bp\sqrt{\kappa+\log(2/\delta)}/\sqrt n$ with probability $\ge1-\delta$, using $B'\le\Bp$, $1\le\Bp$ and $a+b\le\sqrt{2(a^2+b^2)}$. Choosing $n$ so that this is at most $(\log2-\maxfloor)\varepsilon$---that is, $n = O\big((M+1)^2 \Bp^{\,2} (\kappa+\log(2/\delta)) / ((\log2-\maxfloor)^2\varepsilon^2)\big)$---gives $\Pr[\text{misrank under }\hat\theta_n] \le a_{\mathcal{H}}/(\log2-\maxfloor) + \varepsilon$ with probability $\ge1-\delta$. No property of the population maximizer is invoked: the bound is on the \emph{empirical} maximizer directly, through its excess loss.
	\end{proof}
	
	\begin{proposition}[Collision bound on the approximation gap]\label{prop:collision}
		Suppose $(R,\mathcal{H})$ is saturatedly realizable for $L$ (Definition~\ref{def:saturated}). Then
		\[
		a_{\mathcal{H}} \;\le\; \E_{x\sim p_d}\big[\log\!\big(1 + m^+(x)\big)\big],
		\]
		where $m^+(x) = |R(x)\cap L|$. In particular, if $R$ is \emph{orbit-pure} for $L$ almost everywhere---$R(x)\cap L = \emptyset$ for $p_d$-a.e.\ $x$---then $a_{\mathcal{H}} = 0$, and the misranking bound of Theorem~\ref{thm:main} collapses to $\varepsilon$.
	\end{proposition}
	
	\begin{proof}
		Evaluate $-J$ at the saturating scorer $\theta^\star$. For $p_d$-a.e.\ $x$, $s_{\theta^\star}(x)=B$, every $x'\in R^-(x)$ scores exactly $-B$, and every $x'\in R^+(x)$ scores at most $B$. Hence
		\[
		\ell(\theta^\star;x) \;\le\; -B + \log\!\big(e^{B} + m^+(x)\,e^{B} + m^-(x)\,e^{-B}\big)
		\;=\; \log\!\big(1 + m^+(x) + m^-(x)\,e^{-2B}\big).
		\]
		Since $(1+a)(1+b) \ge 1+a+b$ for $a,b\ge0$, and $m^-(x) \le m(x)$,
		\[
		\begin{aligned}\log\!\big(1 + m^+(x) + m^-(x) e^{-2B}\big) &\;\le\; \log\!\big(1+m^+(x)\big) + \log\!\big(1+m(x)e^{-2B}\big)\\ &\;=\; \log\!\big(1+m^+(x)\big) + \ell_0(x).\end{aligned}
		\]
		Taking expectations, $\inf_\theta(-J(\theta)) \le -J(\theta^\star) \le \E[\log(1+m^+(x))] + \meanfloor$, and subtracting $\meanfloor$ gives the bound. Under a.e.\ orbit purity, $m^+(x) = 0$ a.e., so $a_{\mathcal{H}} \le 0$; combined with $a_{\mathcal{H}} \ge 0$ (Lemma~\ref{lem:range}), $a_{\mathcal{H}} = 0$.
	\end{proof}
	
	\begin{remark}[Collisions are intrinsic, not proof slack]\label{rem:intrinsic}
		The collision term cannot in general be removed by a better scorer. If $x' \in R(x)\cap\supp(p_d)$ with both $x$ and $x'$ carrying positive mass and $R(x')\neq\emptyset$ (as boundary support at $x'$ guarantees), the per-example floors at $x$ and at $x'$ make contradictory demands on the single value $s_\theta(x')$: attaining $\ell_0(x)$ requires $s_\theta(x') = -B$, while attaining $\ell_0(x')$ requires $s_\theta(x') = B$. For each fixed value of $s_\theta(x')\in[-B,B]$, at least one of the two per-example gaps $\ell(\theta;\cdot)-\ell_0(\cdot)$ is bounded away from zero, and minimizing their (mass-weighted) sum over the compact range shows $a_{\mathcal{H}} > 0$ strictly whenever, with positive $p_d$-probability, $R(x)$ contains a support point $x'$ with $R(x')\neq\emptyset$. (If $R(x')=\emptyset$, the loss at $x'$ is identically $0$ and places no demand on $s_\theta(x')$.) This is the analogue, for self-generated structured negatives, of the class-collision phenomenon in contrastive representation learning \citep{arora2019}: a perturbation that is itself a valid positive is an inherently ambiguous training signal. Under saturated realizability, the misranking bound of Theorem~\ref{thm:main} is accordingly guaranteed to be non-vacuous when $\E[\log(1+m^+(x))] < \log 2 - \maxfloor$. Since $\log(1+m^+(x))\ge\log 2$ whenever $m^+(x)\ge1$, this requires $\Pr[m^+(x)\ge1]<1-\maxfloor/\log 2$: most positives must have \emph{no} in-language perturbation at all, which is much stronger than a small in-language fraction of each orbit; choosing $R$ to make its perturbations \emph{fail} to be grammatical is part of the operator's job. Appendix~\ref{app:example} exhibits both regimes on a single target language.
	\end{remark}
	
	\begin{remark}[Why $\varepsilon^{-2}$, and why no surrogate-calibration transform]\label{rem:rate}
		The rate is $\varepsilon^{-2}$, the agnostic uniform-convergence rate, and it comes out without a Fisher-consistency/$\psi$-transform argument. The reason is the special geometry of the contrastive loss exploited in Lemma~\ref{lem:cal}: misranking forces $p_0\le\tfrac12$ and hence a definite loss increment $\log2-\maxfloor$ above the per-example floor, which makes the misranking indicator \emph{pointwise} dominated by a linear function of $(\ell-\ell_0(x))$. Generic multiclass surrogate-calibration arguments \citep{tewaribartlett2007,zhang2004mc} relate excess $0/1$ risk to excess surrogate risk through a concave $\psi(u)=O(\sqrt u)$ transform; routed that way the rate would degrade to $\varepsilon^{-4}$, and the $B>\tfrac12\log M$ condition would be unnecessary. We prefer the direct linear bound: it is self-contained, gives the better rate, and exposes the role of the score bound. The condition $B>\tfrac12\log M$ is exactly what makes $\maxfloor<\log2$, i.e.\ what guarantees that at the worst-case loss floor the softmax mass on the positive exceeds $1/2$: it says the substrate is not so capacity-starved \emph{relative to the orbit size $M$} that even its best scorer cannot confidently pick the positive out of the $M{+}1$ candidates $\{x\}\cup R(x)$. A larger $B$ lowers $\maxfloor$ toward $0$ and widens the margin $\log2-\maxfloor$ in the denominator; the condition is mild, requiring only $B$ to grow logarithmically with $M$. Under exact realizability the rate may sharpen further to $\varepsilon^{-1}$ by optimistic-rate (realizable PAC) arguments; whether it does for the contrastive loss is open (Section~\ref{sec:open}).
	\end{remark}
	
	\begin{remark}[Ranking versus a global classifier]\label{rem:rankingform}
		Theorem~\ref{thm:main} is a \emph{conditional} guarantee: each positive is ranked above the members of \emph{its own} orbit. It does not by itself produce a single global threshold $\tau$ separating $L$ from $\Sstar\setminus L$, because score distributions of different orbits may overlap---a string correctly ranked below its own grammatical source may still outscore the source of another orbit. The conditional statement is what the contrastive objective certifies, and we state only that. The global classifier is recovered under an additional assumption in Corollary~\ref{cor:classification}.
	\end{remark}
	
	\begin{corollary}[Global membership classifier under uniform separation]\label{cor:classification}
		Assume, in addition to the hypotheses of Theorem~\ref{thm:main}, the following \emph{uniform separation} condition: there exist $\gamma>0$ and $\eta>0$ such that every near-optimal scorer separates, i.e.\ for each $\theta$ with $-J(\theta) \le \meanfloor + a_{\mathcal{H}} + \eta$ there is a threshold $\tau_\theta$ with $s_{\theta}(x)\ge\tau_\theta+\gamma$ for $p_d$-a.e.\ $x\in L$ and $s_{\theta}(x')\le\tau_\theta-\gamma$ for every $x'\in\mathrm{Cl}_R(\supp p_d)\setminus L$. Train $\hat\theta_n$ with $n = O\big((M+1)^2 \Bp^{\,2}(\kappa+\log(4/\delta))/\eta^2\big)$, and calibrate the threshold
		\[
		\hat\tau \;:=\; \min_{1\le i\le N} s_{\hat\theta_n}(z_i) \;-\; \frac{\gamma}{2}
		\]
		from an independent held-out positive sample $z_1,\dots,z_N \sim p_d$ of size $N = \lceil \varepsilon^{-1}\log(2/\delta) \rceil$. Then with probability at least $1-\delta$, the classifier $h(x)=\indic[s_{\hat\theta_n}(x)>\hat\tau]$ satisfies $h(x') = 0$ for \emph{every} $x'\in\mathrm{Cl}_R(\supp p_d)\setminus L$ and $\Pr_{x\sim p_d}[h(x) = 0] \le \varepsilon$.
	\end{corollary}
	
	The proof is given in Appendix~\ref{app:proofs}. Two honest notes on the assumption. First, separation is asserted over the whole $\eta$-ball of near-optimal scorers, not for a single ideal $\theta^\dagger$: excess-\emph{loss} control (Lemma~\ref{lem:uc}) does not control the function distance between $s_{\hat\theta_n}$ and any particular realizer, so a condition on one scorer cannot be transferred to the learned one without further convexity or identifiability structure; quantifying over the $\eta$-ball is the assumption actually used. Second, the threshold is calibrated from positives alone---no labelled negatives exist in this setting---which is why the margin, rather than a two-sided sample, controls the negative side.
	
	\section{Where the Bias Lives}\label{sec:bias}
	
	A learnability theorem that escapes Gold's negative result must locate its inductive bias precisely. There are exactly three places it can live here, and they do different work.
	
	\paragraph{The operator $R$.} Fixed in advance, applied uniformly, treated as given. It constrains which structural \emph{operations} are informative, not which languages are possible---it is the content of the boundary-support clause, asserting that $R$ exposes the boundary of $L$ almost everywhere. This is the one substantive structural commitment of the construction. Proposition~\ref{prop:collision} sharpens what ``exposes the boundary'' demands: it is not enough that some orbit elements leave $L$; the operator earns its keep to the degree that most positives have \emph{no} in-language perturbation, since every in-language output inflates the collision bound of Proposition~\ref{prop:collision} and, when it lies in the support with a non-empty orbit of its own, the gap itself (Remark~\ref{rem:intrinsic}).
	
	\paragraph{The capacity bound $\kappa$.} What makes the sample complexity polynomial. It is a parameter, not a stipulation: the theorem holds for any finite $\kappa$, with $n$ scaling polynomially in it.
	
	\paragraph{Realizability, collisions, and the approximation gap.} The realizability clause is the standard PAC realizability assumption transposed to the contrastive setting: the target ranking lies in $\mathcal{H}$. It does not hand the learner $\theta^*$; the learner must still find a good $\theta$, which is the content of Lemma~\ref{lem:uc}. Its quantitative residue is the gap $a_{\mathcal{H}} = \inf_\theta(-J(\theta)) - \meanfloor \ge 0$, which reflects two sources: the \emph{substrate's} inadequacy (can $\mathcal{H}$ represent the separation at all?) and the \emph{operator's} collisions (how much of the orbit lies inside $L$?). Proposition~\ref{prop:collision} bounds it when the first source is absent: under saturated realizability, $a_{\mathcal{H}} \le \E[\log(1+m^+(x))]$, and under saturated realizability together with a.e.\ orbit purity, $a_{\mathcal{H}}=0$. The clause should not be oversold: in its uniform form (Definition~\ref{def:uniform}) it is asserted for \emph{every} $L\in\mathcal{C}$ with one fixed $(R,\mathcal{H})$, a strong joint-adequacy claim about the substrate and the class; the theorem itself needs only its pointwise instance for the target at hand. The result has an agnostic form that drops it: without clause~2 of Definition~\ref{def:faithful}, Theorem~\ref{thm:main} still holds verbatim, with $a_{\mathcal{H}}$ now interpreted as the irreducible approximation-plus-collision error of $(\mathcal H, R)$ against the best achievable separation; the misranking bound $a_{\mathcal{H}}/(\log2-\maxfloor)+\varepsilon$ degrades gracefully with the substrate's inadequacy rather than presupposing adequacy. The margin condition of Corollary~\ref{cor:classification} is the additional condition needed to lift ranking to global classification.
	
	\paragraph{The closure criterion.} The criterion in Theorem~\ref{thm:main} is one-step conditional ranking on $\supp(p_d)$; under its margin condition, Corollary~\ref{cor:classification} extends this to rejection of every ill-formed string in $\mathrm{Cl}_R(\supp(p_d))$. Both are weaker than identification of $L$. Neither certifies anything about $L\setminus\mathrm{Cl}_R(\supp(p_d))$, and the one-step criterion does not by itself control the multi-step closure: an operator whose one-step outputs all contain a \emph{local} violation can be ranked perfectly by a local scorer that nevertheless accepts closure elements carrying only long-distance violations. We state this rather than eliding it. Whether natural target classes are reached by the closure of a fixed $R$, with separation on it, is the open question of Section~\ref{sec:open}. This is the same boundary that separates ``fits the closure of the sample'' from ``identifies the target'' \citep{kodner2022,yang2022}; the right response is to name the criterion exactly, which the theorem does.
	
	\section{From Discrimination to Production}\label{sec:production}
	
	The construction is discriminative---it learns to rank---but it is not confined to recognition. The two ingredients the learner already has, the operator $R$ and the trained scorer $s_{\hat\theta}$, are exactly those a Metropolis--Hastings sampler requires.
	
	\begin{definition}[Symmetrized proposal and induced generator]\label{def:generator}
		Write $R^{-1}(x)=\{y:x\in R(y)\}$ for the reverse moves and $\bar R(x)=R(x)\cup R^{-1}(x)$ for the symmetrized neighbourhood, so that $x'\in\bar R(x)\iff x\in\bar R(x')$ by construction. Fix $s := s_{\hat\theta}$ and a seed $x_0$, and let $\mathcal{G}(x_0)$ be the connected component of $x_0$ in the graph with edges $\{(x,x'):x'\in\bar R(x)\}$. On $\mathcal{G}(x_0)$, define a \emph{lazy} Markov chain: from $x$, with probability $\tfrac12$ remain at $x$; otherwise propose $x'\in\bar R(x)$ with probability $q(x'\mid x)=1/|\bar R(x)|$ and accept with probability
		\[
		a(x\to x') = \min\!\Big\{1,\ \exp\!\big(s(x')-s(x)\big)\,\tfrac{q(x\mid x')}{q(x'\mid x)}\Big\},
		\]
		remaining at $x$ on rejection. We assume $|\bar R(x)|<\infty$ for every $x\in\mathcal{G}(x_0)$, so that $q$ is well defined; no uniform bound is needed. Call this the generator induced by $(R,s)$.
	\end{definition}
	
	\begin{proposition}[Normalization-free generator]\label{prop:generator}
		The chain of Definition~\ref{def:generator} is reversible with respect to the measure $\mu(x) = e^{s(x)}$ on $\mathcal{G}(x_0)$, and is irreducible and aperiodic on $\mathcal{G}(x_0)$ by construction. Then: (i) if $\mathcal{G}(x_0)$ is \emph{finite}, the chain has unique stationary law $\pi(x)\propto\exp(s(x))$ on $\mathcal{G}(x_0)$ and converges to it from every initial state; if $\mathcal{G}(x_0)$ is infinite, no stationary probability law exists ($\mu(x) \ge e^{-B}$ for every $x$ makes $\mu$ non-normalizable; see the proof), and the chain is then a reversible local explorer with respect to the $\sigma$-finite measure $\mu$; (ii) the acceptance rule uses only score \emph{differences} and proposal \emph{ratios}, so no global partition function over $\Sstar$ is computed; (iii) every trajectory remains in $\mathcal{G}(x_0)$, so the generator produces only structures reachable from the seed by symmetrized recursive composition.
	\end{proposition}
	
	The proof is in Appendix~\ref{app:proofs}.
	
	\begin{remark}[Reversibility, components, and mixing]\label{rem:generator}
		Three caveats that Definition~\ref{def:R} does not resolve on its own. (i)~\emph{Reversibility is not automatic.} Substitution operators are self-inverse (replacing a constituent by a like-typed one is undone by the reverse replacement) and need no symmetrization; but deletion is not invertible by deletion---a deleted constituent is recovered only by the matching insertion---so the reverse moves $R^{-1}$ must be adjoined, which is exactly what $\bar R$ does. The proposition is stated for $\bar R$ precisely so that the construction is valid for non-self-inverse operators. (ii)~\emph{Components.} $\mathrm{Cl}_{\bar R}(\supp(p_d))$ may decompose into several components; the generator explores the one containing the seed, which is the right semantics for ``generate from $x_0$'' but does not by itself reach other components. (iii)~\emph{Mixing.} We give no mixing-time bound. For a practical generator the chain must mix in time polynomial in the structure size, which depends on the score landscape $s$ and the connectivity of $\bar R$; we leave this open.
	\end{remark}
	
	The same operator that manufactures negatives for learning manufactures candidates for production, and the same scorer that ranks them in comprehension reweights them in generation. The generator computes no global normalizing constant and makes no claim of infinite cardinality: its reach is the component of a seed, and a stationary sampling law exists exactly when that component is finite. On infinite components the chain remains a well-defined local explorer, but normalizable stationarity---like exact identification---is out of reach, which is the generative face of the closure criterion. ``Discriminative'' here therefore does not entail ``recognition-only.''
	
	\section{A Worked Instance and Two Correspondences}\label{sec:instances}
	
	The hypotheses of Theorem~\ref{thm:main} are non-vacuous. Appendix~\ref{app:example} verifies them end to end, with an explicit operator, scorer class and capacity bound and $a_{\mathcal{H}}=0$, on a strictly $2$-local target, together with a cautionary collision-heavy variant on the same target. The two subsections below sketch how the framework corresponds to classical positive-data results. They are correspondences, not verified instantiations: a full instantiation would require establishing boundary support, a capacity bound, and (saturated) realizability for the stated operator and scorer class, which we do not do here.
	
	\subsection{Substitutable context-free languages}
	
	\citet{clarkeyraud2007} prove that the substitutable context-free languages are identifiable in the limit from positive data; $L$ is substitutable if $xuy,\,xvy,\,zuw\in L$ imply $zvw\in L$. In the present setting, $R(x)$ would substitute a substring of $x$ by a string from a fixed inventory (for instance one estimated on data independent of the training sample; Remark~\ref{rem:constituency}), with $\mathcal{H}$ a class of scorers sensitive to substitution contexts. Substitutability alone does not give boundary support, and we do not establish the capacity or realizability clauses for any specific class. Substitutability cuts both ways here, and the collision analysis of Proposition~\ref{prop:collision} is what makes the accounting honest: substitutions across \emph{matching} contexts stay in $L$ by hypothesis---these are exactly the collisions $m^+(x)$---while substitutions across non-matching contexts supply the out-of-language negatives. The instance is therefore governed by the collision-quantified form of the theorem rather than the orbit-pure case: via Proposition~\ref{prop:collision} the bound is non-vacuous only if most positives admit \emph{no} grammatical substitution at all (Remark~\ref{rem:intrinsic}), which is not the typical regime for unrestricted substitution; this is the case the collision-aware refinement of Section~\ref{sec:open} targets. The co-induction of constituency (Remark~\ref{rem:constituency}) is intended to correspond to the Clark--Eyraud congruence learned distributionally.
	
	\subsection{Subregular classes}
	
	The Strictly Local, Strictly Piecewise, and Tier-Based Strictly Local classes are identifiable from positive data \citep{heinz2010,heinz2018,jardineheinz2016,heinzidsardi2013}. Here $R(x)$ would be generated by inserting, deleting, or substituting individual segments (or segments on the relevant tier), with $\mathcal{H}$ a family of factor-counting scorers over the corresponding factors. \citet{hayashikamigaito2025} show these classes are \emph{finitely observable}: membership is fixed by finitely many primitive predicates (factors, subsequences, or tier factors), which yields linear separability over a predicate embedding. For a language defined by a finite set of forbidden factors, subsequences or tier factors, a clipped linear scorer in the corresponding counts (taken on the boundary-padded string) is $B$ on the language and $-B$ off it, as in Appendix~\ref{app:example}; this plays the role of the realizability clause and connects the identification-in-the-limit results to the PAC form used here. Generic segment edits are not orbit-pure (an edit can land back in $L$); Appendix~\ref{app:example} shows how to choose the operator on a strictly $2$-local target so that purity holds identically, and what goes wrong, quantitatively, when it does not.
	
	\section{Related Work}\label{sec:related}
	
	\paragraph{Implicit negatives and the size principle.} The Bayesian size principle \citep{tenenbaumgriffiths2001}---expected-but-absent forms function as implicit negative evidence---is the generative analogue of the present construction: where the size principle uses a generative model to compute expected frequencies and treats absences as informative, the contrastive construction uses $R$ to construct expected perturbations and treats their dispreference as informative. Both make positive-only learning tractable through internal access to expected-but-unobserved structure; the discriminative version avoids the global normalization the generative version requires (Proposition~\ref{prop:generator}).
	
	\paragraph{Statistical theory of contrastive estimation.} The bound in Lemma~\ref{lem:uc} sits in the generalization theory of contrastive learning, which begins with \citet{arora2019}: they analyse downstream classification quality under a latent-class model with i.i.d.\ negatives drawn from the marginal, and \citet{tosh2021} extend the linear-model analysis under multi-view redundancy. Our setting differs on exactly the point these analyses idealize away: the negatives in $R(x)$ are deterministic functions of the positive, so they are \emph{dependent}, which blocks a naive i.i.d.\ analysis and is what forces the conditional (within-orbit) formulation here. The collision term of Proposition~\ref{prop:collision} is the structured-negatives analogue of the class-collision penalty in \citet{arora2019}. \citet{vanelstghoshdastidar2025} develop PAC-Bayesian risk certificates for contrastive objectives that hold under a related form of dependence (SimCLR's reuse of positives as negatives); their refinements indicate that accounting for the dependence structure of $R(x)$ directly, rather than through the conservative vector contraction of Remark~\ref{rem:Mfactor}, is the natural route to sharpening the $(M+1)$ factor.
	
	\paragraph{Relation to active (query) learning.} The classical positive results on identifying language classes are \emph{active}: Angluin's $L^*$ \citep{angluin1987} learns any regular set in polynomial time from a minimally adequate teacher answering membership queries (``is $w\in L$?'') and equivalence queries (returning a counterexample). The present construction is passive and oracle-free---it sees only positive examples and manufactures its own contrasts via $R$. What replaces the membership oracle is the pairing of the operator (which proposes candidates) with boundary support (which guarantees the candidates straddle the boundary often enough to be informative); what replaces the equivalence oracle is nothing, since there is no counterexample feedback, only the contrastive signal. The conditional-ranking criterion, weaker than exact identification, is the price of dispensing with both oracles---and it is the same trade visible in the closure-versus-target gap of Section~\ref{sec:bias}.
	
	\section{Open Problems}\label{sec:open}
	
	\subsection{The mildly context-sensitive conjecture}
	
	The central open question is which class $\mathcal{C}$, in the linguistically relevant range, satisfies the hypotheses with a fixed $R$.
	
	\begin{conjecture}[Bounded-fan-out MCS]\label{conj:mcs}
		For any class $\mathcal{C}$ in the mildly context-sensitive hierarchy whose closure under a fixed multi-component operator $R$ of bounded fan-out is polynomially bounded in input size, there is a hypothesis class $\mathcal{H}$, with capacity $\kappa$ polynomial in the relevant complexity parameter (fan-out for MCFGs, adjunct depth for TAGs, phase depth for minimalist grammars), with the following property. For every $L\in\mathcal{C}$ and every admissible input distribution $p_d$ with support orbit bound $M\le M_0$, the pair $(R_{\mathrm{train}},\mathcal{H})$ is saturatedly realizable and $\E[\log(1+m^+(x))]\le c$ for a constant $c<\log2-\log(1+M_0e^{-2B})$. Here $R_{\mathrm{train}}(x)\subseteq R(x)$ collects the dependency-breaking moves of $R$ (perturbations of one end of a linked tuple); the dependency-preserving co-substitutions are reserved for the closure and the generator, as discussed below. Theorem~\ref{thm:main}, applied to $R_{\mathrm{train}}$, then gives a non-vacuous conditional-ranking guarantee uniformly over $\mathcal{C}$.
	\end{conjecture}
	
	The Clark--Eyraud and subregular results are classical positive-data results at the lower end; only the strictly local instance of Appendix~\ref{app:example} is verified in the present framework. Extending them upward requires a fixed $R$ handling adjoining (TAG), multi-component substitution (MCFG), and phase-based composition (minimalist grammars). Cross-serial dependencies \citep{shieber1985} and long-distance agreement are the diagnostic, but the failure they expose must be located carefully. Perturbing one end of an agreement relation while fixing the other yields an out-of-language string, so boundary support \emph{holds}; what can fail is realizability, when the scorer cannot represent the link. For additive (or clipped-additive) window-based scorers it does fail in the following situation \citep{sengupta2026nonlocal}. Let $a,b$ be the two values of the linked feature and $w$ an intervening string longer than the window. If $awa$ and $bwb$ both carry positive mass and their orbits contain the perturbations of the same end, $awb$ and $bwa$, then the two ranking constraints require the same difference of window scores to be both positive and negative. Operators that avoid exposing the dependency in this way can restore realizability, but may then certify a different, local rule (Section~\ref{sec:bias}). Co-substitution that varies the linked sites \emph{as a unit}---the tuple structure of the MCS formalisms---plays the complementary role: it preserves the dependency, so its outputs are typically in-language. They are collisions for training, but they are exactly the moves the closure $\mathrm{Cl}_R$ and the generator of Section~\ref{sec:production} need in order to reach well-formed strings carrying the dependency.
	
	\subsection{Bounded fan-out and what it bets on}
	
	The word ``bounded'' carries the weight. A fixed fan-out $k$ means $R$ perturbs at most $k$ linked sites at once, for a constant $k$ independent of the input; ``linked'' is defined over the derivation, not surface distance, so the $k$ sites are adjacent in the derivation tree however far apart in the string. A fixed-$k$ operator therefore remains local in the only sense the capacity argument needs while reaching across arbitrary surface spans, and it is universal in the sense of Definition~\ref{def:R}: one operator, applied to every input. Unboundedness lives in how many fixed-$k$ operations compose, not in the fan-out of any one. The empirical bet of the program, in precise form, is that human language sits within some fixed small fan-out---a constant $k$ such that a fan-out-$k$ operator, paired with a bounded-capacity scorer that can represent the dependency, exposes and separates every attested dependency a.e.\ This is Joshi's mildly-context-sensitive hypothesis \citep{joshi1985,stabler1997} read as a claim about contrast operators rather than grammars. It would be refuted by an attested dependency requiring unbounded fan-out to expose; we know of no clear case, but the matter is empirical, not settled by the theorem.
	
	\subsection{Sharper rates and collision-aware calibration}
	
	Three quantitative questions remain. First, the $(M+1)$ factor of Lemma~\ref{lem:uc} is conservative; the $\ell_\infty$ vector contraction and class-count-optimal multiclass bounds should reduce it toward $\sqrt{M+1}$, and a PAC-Bayesian analysis exploiting the dependence structure of $R(x)$ \citep{vanelstghoshdastidar2025} may do better. Second, under exact realizability the $\varepsilon^{-2}$ rate of Theorem~\ref{thm:main} may sharpen to $\varepsilon^{-1}$ via optimistic-rate arguments; whether it does for the contrastive loss is open. Third, the calibration of Lemma~\ref{lem:cal} charges the full collision term to the misranking bound; a refined analysis separating within-orbit collisions from genuine misrankings---for instance by conditioning the loss on the out-of-language sub-orbit, in the spirit of the class-collision decomposition of \citet{arora2019}---could remove the $\E[\log(1+m^+(x))]$ term for collision-heavy operators such as constituent substitution, where it currently dominates the bound.
	
	\section{Conclusion}
	
	We have shown that a learner equipped with a fixed recursive contrast operator and a bounded-capacity discriminative scorer can identify a target language from positive data alone, in a precise one-step conditional-ranking sense, with polynomial sample complexity, whenever the substrate saturatedly realizes the separation and the operator is orbit-pure; otherwise the guarantee holds up to an explicit approximation-gap term---negatives constructed by the learner rather than supplied by an oracle, and no global normalizing constant computed. The misranking bound follows from the geometry of the contrastive loss directly, giving an $\varepsilon^{-2}$ rate, and lifts to a global classifier under a margin condition. The same two ingredients yield a normalization-free generator, so the construction is generative in a controlled sense as well as discriminative. The bias has not vanished but moved: into the operator, the capacity bound, and a realizability clause whose residue---an explicit approximation gap---reflects substrate inadequacy and orbit collisions; under saturated realizability it is bounded by the collision term, and it vanishes when in addition the operator is orbit-pure. Whether a fixed bounded-fan-out operator suffices for the mildly context-sensitive range is the open conjecture the construction bequeaths.

\acks{I thank Halil Talha Ertan for pointing out that the orbit bound in the first version was stated over all of $\Sstar$, which the worked example of Appendix~\ref{app:example} violates; the bound is now restricted to the support of the input distribution. Funding: None. No competing/conflicting interests to declare.}

\newpage

\appendix

\section{Deferred Proofs}\label{app:proofs}

\begin{proof}[Proof of Lemma~\ref{lem:range}]
	Write $u_0 = s_\theta(x)$ and $u_j = s_\theta(x'_j)$, $j=1,\dots,m$ with $m=m(x)$, so $\ell = -u_0 + \log\sum_{k} e^{u_k}$ over the box $[-B,B]^{m+1}$. The map is convex (a log-sum-exp minus a linear term) and $\partial\ell/\partial u_0 = p_0 - 1 \le 0$, $\partial\ell/\partial u_j = p_j \ge 0$, where $p=\mathrm{softmax}(u)$; so $\ell$ is nonincreasing in $u_0$ and nondecreasing in each $u_j$. The minimum over the box is therefore at $u_0=B$, $u_j=-B$, giving
	\[
	\ell_{\min}(x) = -B + \log\!\big(e^{B} + m\, e^{-B}\big) = \log\!\big(1 + m\, e^{-2B}\big) = \ell_0(x),
	\]
	and it is attained only there since $\ell$ is strictly monotone in each coordinate over the interior. The maximum is at $u_0=-B$, $u_j=B$, giving $B + \log(e^{-B} + m\, e^{B}) = 2B + \log(m + e^{-2B}) \le 2B + \log(m+1)$, which is at most $2B+\log(M+1)$ for $x\in\supp(p_d)$. Both extrema are attained because $\ell$ is monotone in each coordinate over the box. Since $\ell_0(x)$ is increasing in $m$ and $m(x) \le M$ on $\supp(p_d)$, we have $\ell_0(x) \le \maxfloor$ there, with equality iff $m(x) = M$; the per-example floor, not $\maxfloor$, is the valid pointwise lower bound when orbit sizes vary. Finally, $-J(\theta) = \E[\ell(\theta;x)] \ge \E[\ell_0(x)] = \meanfloor$, so $a_{\mathcal{H}} \ge 0$.
\end{proof}

\begin{proof}[Proof of Lemma~\ref{lem:uc}]
	The per-example loss is $\ell(\theta;x) = \phi_x\big(s_\theta(x), s_\theta(x'_1),\dots,s_\theta(x'_{m(x)})\big)$, where $\phi_x(u) = -u_0 + \log\sum_k e^{u_k}$ is the multinomial-logistic loss over the $m(x)+1$ active coordinates. By Lemma~\ref{lem:range} the loss at every $x\in\supp(p_d)$ lies in an interval of width at most $B' := 2B+\log(M+1)$. The gradient of $\phi_x$ is $\nabla\phi_x = p - e_0$ with $p$ the softmax over the active coordinates and $e_0$ the indicator of the positive coordinate, so $\|\nabla\phi_x\|_2^2 = (1-p_0)^2 + \sum_{j\ge1}p_j^2 \le (1-p_0)^2 + \big(\sum_{j\ge1}p_j\big)^2 = 2(1-p_0)^2 \le 2$; hence each $\phi_x$ is Lipschitz in the $\ell_2$ norm of its argument with constant $L = \sqrt2$, uniformly in $x$.
	
	The loss class is the composition of the example-dependent Lipschitz functions $\phi_x$ with the tuple of at most $M+1$ evaluations (the sample points lie in $\supp(p_d)$ almost surely, so $m(x_i)\le M$) of a single $s_\theta\in\mathcal{H}$. Embed all tuples in $\R^{M+1}$ by padding the inactive coordinates $j > m(x)$ with a repeat of the positive's own score, $\tau_j(x):=x$ for $j>m(x)$, on which $\phi_x$ does not depend; the Lipschitz constant is unaffected, and every padded evaluation stays in $\mathcal{X}_R$. Each coordinate class is $\{x \mapsto s_\theta(\tau_j(x)) : \theta\in\Theta\}$ for a deterministic map $\tau_j$ ($\tau_0 = \mathrm{id}$, $\tau_j(x) = x'_j$), and its empirical Rademacher complexity on the sample $(x_1,\dots,x_n)$ equals that of $\mathcal{H}$ on the transformed sample $(\tau_j(x_1),\dots,\tau_j(x_n))$, which is at most $\sqrt{\kappa/n}$ by the sample-uniform capacity bound of Definition~\ref{def:hclass}, since the transformed sample lies in $\mathcal{X}_R$---this is where the worst-case (rather than in-expectation) form of the definition is used. By Maurer's vector-contraction inequality \citep{maurer2016}, which permits the Lipschitz function to depend on the example index, the empirical Rademacher complexity of the loss class is at most $\sqrt2\, L \sum_{j=0}^{M} \widehat\Rad_n(\mathcal{H}\circ\tau_j) \le 2(M+1)\sqrt{\kappa/n}$.
	
	Standard symmetrization for bounded losses \citetext{\citealp[Thm.~8]{bartlettmendelson2002}; \citealp[Ch.~3]{mohri2018}} then gives, with probability $\ge 1-\delta$, a uniform deviation $\sup_\theta |\hat J_n(\theta) - J(\theta)|$ bounded by twice this Rademacher term plus a $B'\sqrt{\log(2/\delta)/(2n)}$ concentration term (two-sided, hence $2/\delta$). As empirical maximizer, $\hat\theta_n$ satisfies $-J(\hat\theta_n) \le -\hat J_n(\hat\theta_n) + \sup_\theta|\cdot| \le -\hat J_n(\theta) + \sup_\theta|\cdot| \le -J(\theta) + 2\sup_\theta|\cdot|$ for every $\theta$, which is the stated one-sided excess-risk bound with an absolute constant $C$.
\end{proof}

\begin{proof}[Proof of Proposition~\ref{prop:generator}]
	For $x\ne x'$ with $x'\in\bar R(x)$, the transition kernel is $P(x,x') = \tfrac12\, q(x'\mid x)\, a(x\to x')$, and the symmetry of $\bar R$ gives $q(x'\mid x)>0\iff q(x\mid x')>0$, so the acceptance ratio is well-defined whenever a move is proposed. Then
	\[
	\mu(x)\,P(x,x') = \tfrac12 \min\!\big\{ e^{s(x)} q(x'\mid x),\; e^{s(x')} q(x\mid x') \big\},
	\]
	which is symmetric in $(x,x')$; hence detailed balance $\mu(x)P(x,x') = \mu(x')P(x',x)$ holds and the chain is reversible with respect to $\mu$. The acceptance depends on the target only through the ratio $e^{s(x')-s(x)}$, so no normalizer over $\Sstar$ is computed, giving (ii); and every proposal lies in $\bar R(\cdot)$, so the chain never leaves $\mathcal{G}(x_0)$, giving (iii). Restriction to the component $\mathcal{G}(x_0)$ makes the chain irreducible: every edge $(x,x')$ of the component has proposal probability $q(x'\mid x)=1/|\bar R(x)|>0$ in each direction, and acceptance probability $a(x\to x')\ge\min\{1,\,e^{-2B}|\bar R(x)|/|\bar R(x')|\}>0$, since scores lie in $[-B,B]$ and neighbourhoods are finite. Laziness makes it aperiodic. If $\mathcal{G}(x_0)$ is finite, an irreducible aperiodic chain reversible with respect to $\mu$ has unique stationary law $\pi = \mu/\mu(\mathcal{G}(x_0))$, i.e.\ $\pi\propto e^{s}$, and converges to it from every initial state, giving (i). If $\mathcal{G}(x_0)$ is infinite, then $\mu(\mathcal{G}(x_0)) \ge |\mathcal{G}(x_0)|\, e^{-B} = \infty$, so $\mu$ is $\sigma$-finite but not normalizable. Reversibility makes $\mu$ invariant: $\sum_x\mu(x)P(x,x')=\mu(x')\sum_x P(x',x)=\mu(x')$, the sums being finite by local finiteness. Suppose a stationary probability law $\pi$ existed. An irreducible chain with a stationary probability law is positive recurrent, and the invariant measure of an irreducible recurrent chain is unique up to scaling, so $\pi$ would be proportional to $\mu$, which is impossible since $\mu(\mathcal{G}(x_0))=\infty$. Hence no stationary probability law exists. Whether the chain is recurrent or transient then depends on the geometry of $(\bar R, s)$ and is left open, as discussed in Remark~\ref{rem:generator}.
\end{proof}

\begin{proof}[Proof of Corollary~\ref{cor:classification}]
	Split the failure probability as $\delta/2 + \delta/2$.
	
	\emph{Step 1 (the learned scorer separates).} By Lemma~\ref{lem:uc} with confidence $1-\delta/2$ and the stated $n$, the excess loss of $\hat\theta_n$ is at most $\eta$, i.e.\ $-J(\hat\theta_n) \le \meanfloor + a_{\mathcal{H}} + \eta$. By the uniform separation assumption, there is then a threshold $\tau^\star := \tau_{\hat\theta_n}$ with $s_{\hat\theta_n}(x) \ge \tau^\star + \gamma$ for $p_d$-a.e.\ $x \in L$ and $s_{\hat\theta_n}(x') \le \tau^\star - \gamma$ for every $x' \in \mathrm{Cl}_R(\supp p_d)\setminus L$. Condition on this event.
	
	\emph{Step 2 (no false positives).} The held-out scores satisfy $s_{\hat\theta_n}(z_i) \ge \tau^\star + \gamma$ almost surely, so $\hat\tau = \min_i s_{\hat\theta_n}(z_i) - \gamma/2 \ge \tau^\star + \gamma/2 > \tau^\star - \gamma$. Hence every $x' \in \mathrm{Cl}_R(\supp p_d)\setminus L$ has $s_{\hat\theta_n}(x') \le \tau^\star - \gamma < \hat\tau$ and is classified $h(x') = 0$. This holds pointwise, with no further probability cost.
	
	\emph{Step 3 (few false negatives).} Let $q_\varepsilon := \inf\{t : \Pr_{x\sim p_d}[s_{\hat\theta_n}(x) < t] > \varepsilon\}$ be the lower $\varepsilon$-quantile of the positive score distribution, so that $\Pr[s_{\hat\theta_n}(x) < q_\varepsilon] \le \varepsilon$ and $\Pr[s_{\hat\theta_n}(z) \le q_\varepsilon] \ge \varepsilon$ for a fresh $z\sim p_d$. Then
	\[
	\Pr\Big[\min_{i\le N} s_{\hat\theta_n}(z_i) > q_\varepsilon\Big] \;\le\; (1-\varepsilon)^N \;\le\; e^{-\varepsilon N} \;\le\; \delta/2
	\]
	for $N \ge \varepsilon^{-1}\log(2/\delta)$. On the complementary event, $\min_i s_{\hat\theta_n}(z_i) \le q_\varepsilon$, and since $\hat\tau < \min_i s_{\hat\theta_n}(z_i)$,
	\[
	\Pr_{x\sim p_d}[h(x) = 0] = \Pr\big[s_{\hat\theta_n}(x) \le \hat\tau\big] \le \Pr\big[s_{\hat\theta_n}(x) < \min_i s_{\hat\theta_n}(z_i)\big] \le \Pr\big[s_{\hat\theta_n}(x) < q_\varepsilon\big] \le \varepsilon.
	\]
	A union bound over Steps 1 and 3 completes the proof.
\end{proof}

\section{A Worked Instance, and Its Collision-Heavy Shadow}\label{app:example}

\begin{example}[A verified instance with $a_{\mathcal{H}}=0$]\label{ex:sl2}
	The hypotheses can be checked end-to-end on a concrete case. Let $\Sigma=\{a,b\}$ and let $L$ be the strictly $2$-local language of strings with no factor $bb$. Take the \emph{$b$-doubling} operator
	\[
	R(x) \;=\; \big\{\, x \text{ with its $i$-th occurrence of $b$ doubled} \;:\; 1 \le i \le \#_b(x) \,\big\},
	\]
	so $m(x) = \#_b(x)$, the number of $b$'s in $x$. The operator is universal (defined on every string, with no reference to $L$), local (insertion of a single segment adjacent to an existing one), and recursive (the occurrences of $b$ in $uv$ are those of $u$ together with those of $v$, so doubling within $u$ yields $u'v$ with $u'\in R(u)$, and likewise for $v$).
	
	\emph{Orbit purity holds identically:} every output contains the factor $bb$ at the doubled site, so $R(x)\subseteq\Sstar\setminus L$ for every $x\in L$---not merely almost everywhere. \emph{Boundary support} holds at every $x$ containing at least one $b$; strings in $a^*$ have empty orbits and carry no contrastive signal, so we require $\supp(p_d)\cap a^* = \emptyset$, consistent with the a.e.\ formulation of Definition~\ref{def:faithful}.
	
	Let $\mathcal{H}$ be the bounded affine scorers in bigram-count features, $s_\theta(x)=\mathrm{clip}_{[-B,B]}\big(\theta_0+\sum_{cd\in\Sigma^2}\theta_{cd}\,c_{cd}(x)\big)$, where $c_{cd}(x)$ counts the factor $cd$ in $x$. The parameter $\theta_0=B,\ \theta_{bb}=-2B,\ \theta_{cd}=0$ otherwise gives $s(x)=B$ for every $x\in L$ (no $bb$, so the clip returns $\theta_0=B$) and $s(x')=-B$ for every string containing a $bb$ (the count is $\ge1$, so $\theta_0+\theta_{bb}c_{bb} \le -B$ before clipping)---hence for every element of every orbit. Saturated realizability (Definition~\ref{def:saturated}) therefore holds with this explicit $\theta^\star$, and since $m^+ \equiv 0$, Proposition~\ref{prop:collision} gives $a_{\mathcal{H}}=0$: the realizer attains the per-example floor $\ell_0(x)$ at every $x$, and Theorem~\ref{thm:main} yields conditional misranking probability at most $\varepsilon$. For the capacity, restrict the parameters to $\|(\theta_0,(\theta_{cd})_{cd})\|_2\le\sqrt5\,B$, which contains $\theta^\star$. Under a length bound $K$ on $\supp(p_d)$, every string in $\mathcal{X}_R$ has length at most $K+1$, so its bigram counts sum to at most $K$ and $\|(1,(c_{cd}(x))_{cd})\|_2\le K+1$. The linear class then has empirical Rademacher complexity at most $\sqrt5\,B\,(K+1)/\sqrt n$ on every sample from $\mathcal{X}_R$, and clipping is $1$-Lipschitz, so $\kappa = 5B^2(K+1)^2$ \citep{bartlettmendelson2002,mohri2018}. The orbit size is unbounded over $\Sstar$ ($m((ab)^n)=n$), but on the support $m(x)=\#_b(x)\le\lceil K/2\rceil$, since strings in $L$ contain no factor $bb$, so the support orbit bound of Definition~\ref{def:R} is $M\le\lceil K/2\rceil$ and $B>\tfrac12\log\lceil K/2\rceil$ suffices---a score bound growing only logarithmically in string length, and a capacity growing as $B^2K^2$. This is a minimal witness that boundary support, saturated realizability, orbit purity, and the score bound are simultaneously satisfiable by an explicit bounded-weight scorer.
\end{example}

\begin{example}[The collision-heavy shadow: single-symbol substitution]\label{ex:sl2bad}
	On the same target $L$, consider instead the single-symbol substitution operator $R'(x) = \{$flip one position of $x\}$, which is the generic segment-edit operator of Section~\ref{sec:instances}. It is \emph{not} orbit-pure: $aba \mapsto aaa \in L$. Worse, such collisions typically lie in the support: if $aaa \in \supp(p_d)$, then the per-example floors at $aba$ and at $aaa$ make contradictory demands on the single value $s(aaa)$ ($-B$ versus $+B$), so by Remark~\ref{rem:intrinsic} no scorer attains both and $a_{\mathcal{H}} > 0$ strictly. Concretely, the scorer $\theta^\star$ of Example~\ref{ex:sl2} (which must score $aaa$ at $+B$, since $aaa$ is a positive) incurs per-example loss
	\[
	\ell(\theta^\star; aba) = \log\!\big(2 + 2e^{-2B}\big) \;>\; \log 2 \;>\; \ell_0(aba) = \log\!\big(1+3e^{-2B}\big),
	\]
	a gap that does not shrink as $B$ grows (the last inequality uses $B>\tfrac12\log3$, which the theorem's condition $B>\tfrac12\log M$ implies whenever $aba\in\supp(p_d)$, since then $M\ge m(aba)=3$). The gap is not an artefact of $\theta^\star$. Because $R'$ is symmetric, $aaa\in R'(aba)$ and $aba\in R'(aaa)$; for any $\theta$, with $d=s_\theta(aaa)-s_\theta(aba)$, the two per-example losses are at least $\log(1+e^{d})$ and $\log(1+e^{-d})$, whose sum is at least $2\log 2$. If $p_d$ places equal mass $p$ on each of $aba$ and $aaa$, these two strings alone contribute at least $2p(\log2-\maxfloor)$ to $a_{\mathcal{H}}$. Summing over disjoint, equally weighted pairs of this kind shows that $a_{\mathcal{H}}/(\log2-\maxfloor)$ is at least their total mass, so for input distributions concentrated on such pairs the bound is at least (nearly) one, rendering Theorem~\ref{thm:main} vacuous for $R'$ \emph{even though} realizability in the sense of Definition~\ref{def:faithful} holds. The comparison of the two examples is the practical content of Proposition~\ref{prop:collision}: the choice of operator, not only of hypothesis class, is what the loss geometry constrains, and a well-chosen operator is one whose perturbations reliably exit the language.
\end{example}

\end{document}